\documentclass[letterpaper,submission]{article} % DO NOT CHANGE THIS
\usepackage{aaai25}  % DO NOT CHANGE THIS
\usepackage{times}  % DO NOT CHANGE THIS
\usepackage{helvet}  % DO NOT CHANGE THIS
\usepackage{courier}  % DO NOT CHANGE THIS
\usepackage[hyphens]{url}  % DO NOT CHANGE THIS
\usepackage{graphicx} % DO NOT CHANGE THIS
\usepackage{natbib}  % DO NOT CHANGE THIS AND DO NOT ADD ANY OPTIONS TO IT
\usepackage{caption} 
\usepackage{algorithm}
\usepackage{algorithmic}
\usepackage{comment}  %SMSADD
\usepackage{booktabs}
\usepackage{amsmath}
\usepackage{booktabs}
\usepackage{multirow}
\usepackage{colortbl}
\usepackage{graphicx}
\usepackage{subfigure}
\usepackage{dsfont}
\usepackage[listings, many]{tcolorbox}

\newcommand{\bricc}{\textsc{Bricc}}
\usepackage{newfloat}
\usepackage{listings}
\DeclareCaptionStyle{ruled}{labelfont=normalfont,labelsep=colon,strut=off} % DO NOT CHANGE THIS
\floatstyle{ruled}
\newfloat{listing}{tb}{lst}{}
\floatname{listing}{Listing}

\title{AI-Powered Detection of Inappropriate Language in Medical School Curricula}
\author{
    Chiman Salavati\textsuperscript{\rm 1}, 
    Shannon Song\textsuperscript{\rm 2}, 
    Scott A.\ Hale\textsuperscript{\rm 3,4}, 
    Roberto E.\ Montenegro\textsuperscript{\rm 5},\\
    Shiri Dori-Hacohen\textsuperscript{\rm 1}\thanks{Senior authors.},
    Fabricio Murai\textsuperscript{\rm 2}\footnotemark[1]
}
\affiliations{
    \fontsize{10pt}{12pt}\selectfont
    \textsuperscript{\rm 1}University of Connecticut, Storrs, Connecticut, USA\\ 
    \textsuperscript{\rm 2}Worcester Polytechnic Institute, Worcester, Massachusetts, USA\\ 
    \textsuperscript{\rm 3}Meedan, San Francisco, California, USA\\
    \textsuperscript{\rm 4}University of Oxford, Oxford, UK\\ 
    \textsuperscript{\rm 5}University of Washington, Seattle, Washington, USA \\
    \{chiman.salavati, shiridh\}@uconn.edu, \{smsong, fmurai\}@wpi.edu,\\ scott@meedan.com, roberto.montenegro@seattlechildrens.org
}

\begin{document}

\maketitle

\begin{abstract}
The use of inappropriate language---such as outdated, exclusionary, or non-patient-centered terms---in medical instructional materials can significantly influence clinical training, patient interactions, and health outcomes. Despite their reputability, many materials developed over past decades contain examples now considered inappropriate by current medical standards. Given the volume of curricular content, manually identifying instances of inappropriate use of language (IUL) and its subcategories for systematic review is prohibitively costly and impractical. 
To address this challenge, we conduct a first-in-class evaluation of small language models (SLMs) fine-tuned on labeled data and pre-trained LLMs with in-context learning on a dataset containing approximately 500 documents and over 12,000 pages. 
For SLMs, we consider: (1) a general IUL classifier, (2) subcategory-specific binary classifiers, (3) a multilabel classifier, and (4) a two-stage hierarchical pipeline for general IUL detection followed by multilabel classification. For LLMs, we consider variations of prompts that include subcategory definitions and/or shots.
We found that both LLama-3 8B and 70B, even with carefully curated shots, are largely outperformed by SLMs. While the multilabel classifier performs best on annotated data, supplementing training with unflagged excerpts as negative examples boosts the specific classifiers’ AUC by up to 25\%, making them most effective models for mitigating harmful language in medical curricula.

\end{abstract}

\section{Introduction}

The influential role of language in medical records in shaping clinicians' attitudes and behaviors is well-established in the literature~\cite{park2021physician,p2018words}. Stigmatizing and approving language
%---such as referring to a patient with sickle cell disease as ``narcotic dependent'' or ``in our ED frequently''---
 can transmit bias, influence subsequent clinician perceptions, with seemingly objective records ultimately affecting the quality and fairness of patient care \cite{chapman2013physicians,lindquist2015role,glassberg2013among,himmelstein2022examination,fernandez2021words,beach2021testimonial,sun2022negative}. For example, language choices such as ``substance abuser'' instead of ``having a substance use disorder'' shape stigmatizing attitudes, even among mental health professionals~\cite{kelly2010does,ashford2019abusing}. Patients characterized by advanced age, low health literacy, and obesity are often viewed negatively by healthcare professionals in ways that adversely affect the care they receive \cite{kelly2007physician,berkman2011low,webster2022social}.

\begin{figure}[t!] 
    \centering
    \includegraphics[width=\columnwidth]{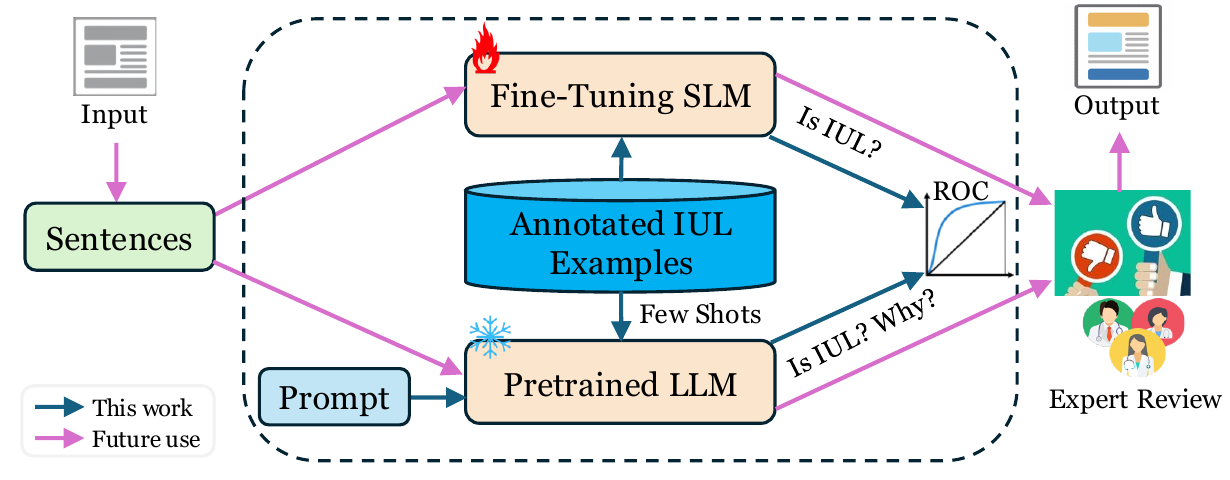}
\caption{IUL detection overview in medical curricula. Dashed line highlights the scope of this work.}
  \label{fig:intro}
\end{figure}

Despite efforts to promote inclusive, patient-centered language, IUL remains prevalent in clinical documentation and interactions \cite{kelly2010does,ashford2019biased,andraka2018qualitative}, often originating from instructional materials and medical training practices \cite{chapman2013physicians}. The field’s reliance on tradition and apprenticeship-based learning perpetuates implicit and explicit biases across generations \cite{fitzgerald2017implicit,madara2020america,salavati2024,butts2024towards}, with IUL in medical education reinforcing harmful narratives and shaping how future healthcare professionals conceptualize care \cite{p2018words}.

In this context, IUL refers to terms or expressions used to describe social identities that are inadequate by current medical standards. IUL focuses on the form of expression—such as implying binaries (e.g., “both males and females”), using outdated or judgmental terms (e.g., “fat,” “fertile,” “mental retardation”), or misusing sex and gender labels—rather than the bias in the claim itself \cite{puhl2013motivating,keyes2010stigma,dickinson2017use,paton2024our}. It can also manifest in exclusive pronouns or the conflation of race and ethnicity \cite{krishnan2019addressing}. While IUL may accompany biased content, it can also occur independently. %Language choices like ``substance abuser'' instead of ``having a substance use disorder'' have been shown to shape stigmatizing attitudes, even among mental health professionals \cite{kelly2010does,ashford2019biased,ashford2019abusing}.
IUL contributes to the perpetuation of stigma, including obesity-related and weight-based stigma, and appears disproportionately in records of non-Hispanic Black patients, raising concerns about exacerbating health disparities \cite{forhan2013inequities,himmelstein2022examination,beach2021testimonial,sun2022negative}.

Despite growing efforts to improve language standards in medical education \cite{p2018words,ashford2019biased,fitzgerald2017implicit}, there is a notable lack of scalable and systematic methods to detect IUL in educational content. Current approaches rely predominantly on manual review by domain experts, which is time consuming and impractical given the vast scale and historical depth of educational materials \cite{salavati2024}. The nuanced nature of IUL, where subtle wording choices can have significant implications, further exacerbates these resource costs. This gap underscores the necessity of developing automated, reliable tools to assist in the comprehensive detection of IUL patterns in medical texts.

% \fm{What is the research gap?} 
In this study, we develop new AI models for detecting IUL patterns in clinical documentation. Building on the Bias Reduction in Curricular Content (BRICC) dataset~\cite{salavati2024}, we implemented and evaluated a series of language models (small and large) to detect IUL in medical texts (see Figure~\ref{fig:intro}), including: (1) a general binary classifier using BioBERT/DistilBERT as backbones, (2) subcategory-specific classifiers for various IUL types, (3) a multi-label classifier to handle samples with multiple IUL subcategories, and (4) a two-stage hierarchical classification pipeline combining general detection with subcategory identification. To our knowledge, this constitutes the first end-to-end AI framework specifically targeting IUL detection in medical educational content. Additionally, for LLMs, we consider
variations of prompts that include subcategory definitions and/or shots. We compared the performance of all models using key evaluation metrics including precision, recall, F1 score, F2 score, and AUC. Comparative analysis against baseline models demonstrates that our proposed approach consistently outperforms baselines across all evaluation criteria. Our contributions are as follows:

\begin{itemize}
    \item We expanded the original labels from the BRICC dataset~\cite{salavati2024} to indicate IUL occurrences through several interactions with the PI who led the collection and labeling of the original dataset.

    \item
    We developed a novel, first-in-class IUL detection system %by adapting and extending existing methodology from \cite{salavati2024}
    for the specialized task of identifying IUL within medical curricula.

    \item We conducted a comprehensive evaluation of language models' performance in detecting IUL in medical texts.

\end{itemize}

\begin{table*}[t]
\renewcommand{\arraystretch}{1.2}
\centering
\footnotesize

\begin{tabular}{p{0.35\linewidth}p{0.3\linewidth}p{0.28\linewidth}}
    \toprule
    \textbf{IUL Subcategory Definition} & \textbf{Example Quote} & \textbf{Annotator Comment} \\
    \midrule
    \textbf{Gender Misuse:} Using gendered terms (e.g., ``women'', ``men'') where anatomical or sex-based references are more appropriate, particularly in population-level statements. This can reinforce gender stereotypes and exclude non-binary individuals. & \textit{``Metformin, which decreases insulin sensitivity, can restore menstrual cycles in 30-50\% of women with PCOS''} & \textit{Consider sex instead of gender, and language that doesn't reinforce sex and gender binaries Men and woman implies gender binary, include other genders/sexes statistics. Use sex terms when speaking of populations.} \\
    % \hline
    \textbf{Sex Misuse:} Refers to the incorrect use of sex terms (e.g., ``male'', ``female'') when referring to individuals rather than biological or population-level characteristics. %Examples include phrases like ``a 49-year-old male'' instead of ``a 49-year-old man''.
    & \textit{``78y/o female presents to primary care with complaint of rash on feet, legs and arms for one month.''} & \textit{Use woman (gender terms) over female (sex terms) in case studies} \\
    % \hline
    \textbf{Age Language Misuse:} Involves the use of vague or stigmatizing age references such as ``young people'' or ``the elderly''. These should be replaced with objective numerical ranges to maintain clarity and avoid stereotyping. & \textit{``Dietary total energy requirements for older adults decline slightly with changes in body composition, metabolic rate, and physical activity.''} & \textit{Consider using an age-range for `older adults'.} \\    
    % \hline
    \textbf{Exclusive Language:} Language that assumes binary sex or gender categories (e.g., ``both males and females'') and excludes individuals outside these binaries. %Inclusive phrasing like ``all patients'' is recommended.
    & \textit{``A woman or man can choose the right method for them and use it to its full potential. The patient’s choice of contraception may differ from the provider’s suggestion. Remember it is up to her, not you.''} & \textit{Consider using an individual over woman and man} \\
    % \hline
    \textbf{Non-patient-centered Language:} Describes individuals primarily by their conditions (e.g., ``diabetics'', ``alcoholics''), rather than as people first. %Preferred alternatives include ``patients with diabetes'' or ``individuals with alcohol use disorder''.
    & \textit{``Diabetics with periodontal disease experience greater difficulty achieving glycemic control.''} & \textit{Non-patient-centered language for `Diabetics' Consider `patients with diabetes} \\
        % \hline
    \textbf{Outdated Term:} Terms that are no longer appropriate in modern medical contexts (e.g., ``mentally retarded'', ``fat and fertile female''). %Such terms should be replaced with respectful and medically accurate language (e.g., ``intellectual disability'', ``patients with higher BMI'').
    & \textit{``The repeats disrupt the function of the FMRP protein, which is involved in synaptic function, which is why the syndrome involves mental retardation.''} & \textit{Mental retardation is an outdated term, consider using alternatives} \\
    \bottomrule
\end{tabular}
\caption{ Examples of IUL across different subcategories. Each row presents a sample quote exhibiting a specific type (subcategory) of IUL—such as gender misuse, sex misuse, age-related bias, exclusive language, non–patient-centered language, or outdated terminology—along with annotator comments suggesting more appropriate alternatives.
}
\label{tab:IUL_Samples}
\end{table*}

\begin{table*}[t]
\renewcommand{\arraystretch}{1.2}
\centering
\footnotesize
\begin{tabular}{p{0.35\linewidth}p{0.3\linewidth}|p{0.28\linewidth}}
    \toprule
    \textbf{Example of Annotated Negative (AN)} & \textbf{Annotator Comment} & \textbf{Example of Extracted Negative (EN)}  \\
    \midrule
    \textbf{(Gender Misuse)} \textit{``Often, significant changes in a child’s growth reflect significant events in the family unit such as a mother going to work, parents separating, moving to a new home or a significant family illness.''} & \textit{This statement reinforces traditional family structures which stigmatizes mothers going to work, or families without a mother. While this reflects gender bias, it does not constitute gender misuse.} & \textit{``He  has no testicular mass but has a small reactive hydrocele.''}\\
    % \hline
    \textbf{(Sex Misuse)} \textit{``Numerous measures of sexual function change as males age, including a decline in the frequency of orgasms, an increase in erectile dysfunction (ED), and a decline in the quality and quantity of sexual thoughts and enjoyment.''} & \textit{suggest citation and review for accuracy, unclear if these should be specified to only males. While this reflects sex bias, it does not constitute sex misuse.} & \textit{``Studies of 46,XY individuals with androgen insensitivity syndrome (AIS) and assigned female  sex at birth have also been revealing.''}\\
    % \hline
    \textbf{(Age Language Misuse)} \textit{``Hereditary pancreatitis (HP) is an autosomal dominant disease with 80\% penetrance, characterized by recurrent episodes of pancreatitis from childhood with a familial occurrence''} & \textit{Childhood captures anyone under 18 yo therefore an appropriate use here.} & \textit{``How does type 1 diabetes present differently in children  and adults?''}\\    
    % \hline
    \textbf{(Exclusive Language)} \textit{``The gross morphological appearance of the nuclear chromatin differs in cells between males and females.} & \textit{cite for sex difference. While this reflects sex bias, it does not constitute exclusive language. } & \textit{``The common  urogenital sinus in a female may be repaired to prevent urinary-tract infections that could lead to  kidney damage.''}\\
    % \hline
    \textbf{(Non-patient-centered Language)} \textit{``A landmark study detailing the clinical features of alcoholic hepatitis; also, one of the first to demonstrate a  potential benefit from corticosteroid therapy.''} & \textit{This does not contain non–patient-centered language because “alcoholic” here refers to the disease alcoholic hepatitis, not to individuals.} & \textit{``Recognize the importance of careful history and physical examination in the evaluation of  cancer patients.''}\\
        % \hline
    \textbf{(Outdated Term)} \textit{``Psychomotor retardation or agitation nearly every day that is observable by others''} & \textit{Correct use of retardation. Great example of "negative sample"} & \textit{``Indication of liver transplantation in severe  alcoholic liver cirrhosis: quantitative evaluation and optimal timing.''}\\

    \bottomrule
\end{tabular}
\caption{Examples of non-IUL (negative) samples across various IUL subcategories. These quotes do not exhibit IUL but occur in contexts typically associated with IUL categories such as gender, sex, or outdated terminology. They serve as hard negative samples that closely resemble IUL instances, making them useful for training and evaluating models to detect subtle IUL.
}
\label{tab:Non_IUL_Samples}
\end{table*}

\section{Related Works}

Through a comprehensive scoping review of academic and gray literature, \citet{healy2022reduce} synthesized core principles and strategies to guide clinicians in avoiding stigmatizing language in medical records, emphasizing language's role in reinforcing or disrupting health disparities. However, these studies primarily focus on clinical documentation and diagnostic labels, without addressing instructional materials as a potential root cause of stigmatizing language. Moreover, they do not propose concrete solutions for identifying and mitigating IUL in educational content used for training healthcare professionals. This gap is especially concerning, as instructional materials are often used to train future clinicians and also serve as foundational data sources for AI systems. IUL embedded in these materials not only reinforces human biases but also risks being learned and amplified by AI models deployed in clinical settings.

% \fm{However, these works neither focus on instructional materials as one of the root causes nor propose a solution...}

% \fm{This is a bit disconnected from the previous paragraph. It refers to training models with biased data. You need to introduce this task first and try to emphasize the IUL as a problem, in addition to the bias.} Researchers have raised growing concerns about the ethical risks and challenges of deploying AI in healthcare without sufficient oversight \cite{gianfrancesco2018potential}. A major source of this concern lies in how training data are collected, as biased or incomplete datasets can cause models to generate unfair or unreliable predictions. Several studies \cite{gianfrancesco2018potential,mittermaier2023bias} highlight how data quality problems in machine learning pipelines may disproportionately benefit certain populations over others. There is increasing advocacy for formal AI governance frameworks to ensure accountability and responsible deployment. \citet{nelson2019bias} highlights the crucial role clinicians must play in overseeing and validating AI systems, while \citet{kiyasseh2023human} emphasize the need for explainable models that allow regulatory bodies, such as the FDA, to establish and enforce effective bias management protocols.

Given the growing integration of AI in healthcare, researchers have raised concerns about the ethical risks and challenges of deploying these systems without sufficient oversight \cite{gianfrancesco2018potential}. A major source of this concern lies in how training data are collected, as biased or incomplete datasets can cause models to generate unfair or unreliable predictions. Several studies \cite{gianfrancesco2018potential,mittermaier2023bias,Olulana2024aies} highlight how data quality problems in machine learning pipelines may disproportionately benefit certain populations over others. There is growing advocacy for formal AI governance frameworks to ensure accountability and responsible deployment. \citet{nelson2019bias} highlights the crucial role clinicians must play in overseeing and validating AI systems, while \citet{kiyasseh2023human} emphasize the need for explainable models that allow regulatory bodies, such as the FDA, to establish and enforce effective bias management protocols. In line with these works, \citet{dori2021fairness} propose a multifaceted framework that integrates perspectives from medical education, sociology, and antiracism to promote fairness in healthcare AI. Building on this foundation, our work leverages AI models to detect IUL in medical text, aligning with the fairness goals articulated by \citet{dori2021fairness}.

\citet{yenala2018deep} has addressed the detection of IUL in user-generated content, such as search queries and online conversations, using deep learning models like the Convolutional Bi-Directional LSTM. 
\citet{jain2023inappropriate} proposed a two-step approach for detecting and rephrasing offensive language using advanced computational linguistic techniques, achieving high classification accuracy while preserving semantic meaning to promote respectful communication. \citet{mishra2024detection} evaluated the effectiveness of machine learning and deep learning models for detecting offensive language on social media, finding that BERT outperformed other approaches in accuracy and F1-score, though with higher computational costs. However, these models are not specialized for medical textual data, where language carries clinical significance and domain-specific nuances.

Conversely, \citet{salavati2024} focuses specifically on bias detection rather than general IUL. The study introduces BRICC, an expert-annotated dataset of medical curricula aimed at detecting medical bisinformation (biased information) \cite{dori2021fairness} that continues to be taught despite being inaccurate. The authors trained and evaluated several AI models to identify and flag medical text with potential bias for expert review. While that work makes significant strides in mitigating bias in medical texts, it does not address the forms of IUL discussed in our study---such as outdated, exclusionary, or non–patient-centered language---which do not fall under the traditional definition of bias but still carry harmful implications.
% \fm{Cite some approaches that try to optimize the model or align the model so as to reduce bias. Then indicate that this approach has shortcomings in that it doesn't generalize that well for new models and that, instead, we build our approach on the Fairness via AI principle...} Expanding these conversations, \citet{dori2021fairness} propose a multifaceted framework that integrates perspectives from medical education, sociology, and antiracism to promote fairness in healthcare AI. Building on this foundation, our work leverages AI models to detect IUL in medical text, aligning with the fairness goals articulated by \citet{dori2021fairness}.  

% \fm{add AI-based approaches for detecting bias or other harmful language in general text, not medical. then mention that these models are not specialized for medical textual data.}

% \fm{add another para: the only exception is AIES2024, but they focus on bias instead.} \fm{talk about AIES24 paper}

% \fm{this should be combined with para that starts with `Researchers...'} There is increasing advocacy for formal AI governance frameworks to ensure accountability and responsible deployment. \citet{nelson2019bias} highlights the crucial role clinicians must play in overseeing and validating AI systems, while \citet{kiyasseh2023human} emphasize the need for explainable models that allow regulatory bodies, such as the FDA, to establish and enforce effective bias management protocols.

\begin{figure*}[htb]
    \centering \includegraphics[width=1\linewidth]{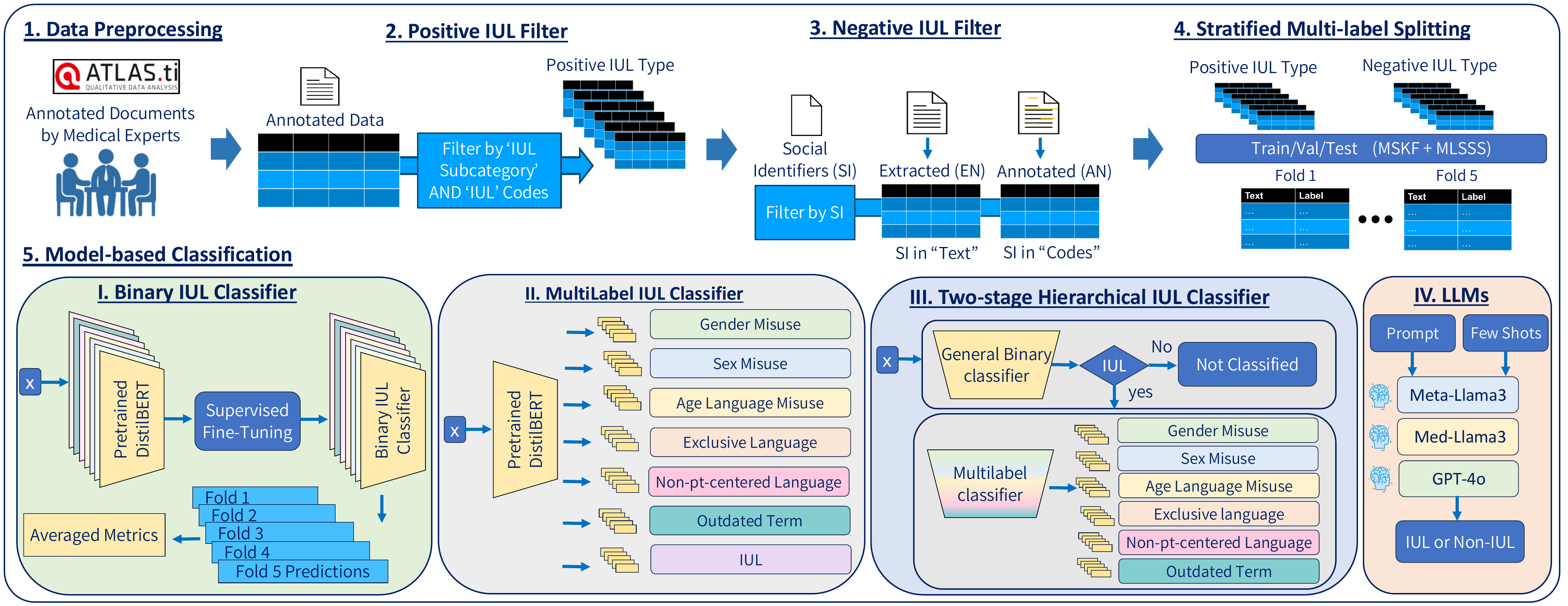}
    \caption{ Overview of our proposed IUL detection pipeline. In Step 1, we preprocess annotated medical documents using expert-labeled data from ATLAS.ti. In Step 2, we filter positive samples by both IUL subcategory and IUL codes. In Step 3, we filter negative samples by detecting social identifiers in text and codes, constructing clean negative sets. In Step 4, we apply a two-stage split strategy to create training, validation, and test partitions for robust evaluation. For Step 5, we implement and evaluate three detection strategies: (I) a binary IUL classifier for general and specific IUL detection, (II) a multilabel classifier for predicting multiple IUL subcategories simultaneously, (III) a two-stage hierarchical classifier that first identifies IUL presence and then predicts specific subcategories for positively flagged samples, and (IV) an experiment of different LLMs with various prompts.}
    \label{fig:overview}
\end{figure*}

\section{Problem Definition}

The overarching objective of this work is to develop an automated, scalable, and robust pipeline that supports human experts in systematically reviewing large-scale collections of medical educational texts. Ultimately, this can accelerate the detection of IUL while upholding the highest standards of quality, inclusivity, and equity in medical education. Therefore, we frame IUL detection as a machine learning task. 

Formally, let \( x \) denote a medical educational text excerpt, which may include clinical claims, case reports, epidemiological statistics, or other instructional content. We define \( y \in \{0,1\} \) as a label indicating the presence (\( y = 1 \)) or absence (\( y = 0 \)) of IUL (in general). For excerpts identified as containing IUL (\( y = 1 \)), we further assign a multilabel vector \( \mathbf{z} = (z_1, z_2, \ldots, z_C) \), where \( C \) is the number of predefined IUL subcategories, and each \( z_c \in \{0,1\} \) denotes whether the excerpt exhibits the \( c \)-th subcategory of IUL.

To illustrate, consider this example shown in Table~\ref{tab:IUL_Samples}:  
\textit{``78y/o female presents to primary care with complaint of rash on feet, legs, and arms for one month.''}  
This excerpt is labeled \( y = 1 \) (IUL present) and \( z_\text{sex misuse} = 1 \), reflecting the inappropriate application of sex terminology where gendered or person-first language would be more appropriate.

We formalize IUL detection as a two-stage supervised learning task:

\subsection{Task 1: General IUL Detection}

Given an input excerpt \( x \), the first task is to learn a binary classification function:
\[
f_\theta: x \mapsto \hat{y} \in \{0,1\},
\]
where \( \theta \) represents the model parameters and \( \hat{y} \) denotes the predicted general IUL label. This stage acts as a broad screening mechanism, identifying excerpts that warrant expert review. As part of our proposal of an expert-in-the-loop framework, we emphasize high recall, accepting some false positives to minimize the risk of missing harmful or inappropriate content.

\subsection{Task 2: IUL Subcategory Detection}

For excerpts identified as containing IUL (\( y = 1 \)), the second task focuses on fine-grained multilabel classification to determine the specific IUL subcategories. We define:
\[
g_\phi: x \mapsto \hat{\mathbf{z}} = (\hat{z}_1, \hat{z}_2, \ldots, \hat{z}_C), \quad \hat{z}_c \in \{0,1\},
\]
where \( \phi \) denotes the model parameters and each \( \hat{z}_c \) indicates the predicted presence of the \( c \)-th IUL subcategory.

\section{Dataset}

We have obtained the BRICC dataset introduced by~\citet{salavati2024}, which comprises over 12,000 pages of medical instructional materials---including syllabi, lecture slides, and assigned readings---collected from the University of Washington School of Medicine used in classes that span two academic years. These materials cover a slew of curricular topics, such as \textit{Lifecycle} and \textit{Mind, Brain, and Behavior}.

The dataset was annotated for bias and IUL occurrences by a team of trained annotators under the supervision of a domain expert using a detailed coding manual. The annotation followed a rigorous process where each excerpt underwent independent review by multiple annotators.

In total, the dataset includes more than 4,000 annotated excerpts capturing various forms of IUL and bias. While annotations encompass a broad range of identity-related misuse types (including race and ethnicity), this study focuses on the six most prevalent categories: \textit{gender misuse}, \textit{sex misuse}, \textit{age-related language misuse}, \textit{exclusive language}, \textit{non--patient-centered language}, and \textit{outdated terminology}.

\begin{figure}[ht] 
    \centering
    \includegraphics[width=0.9\columnwidth]{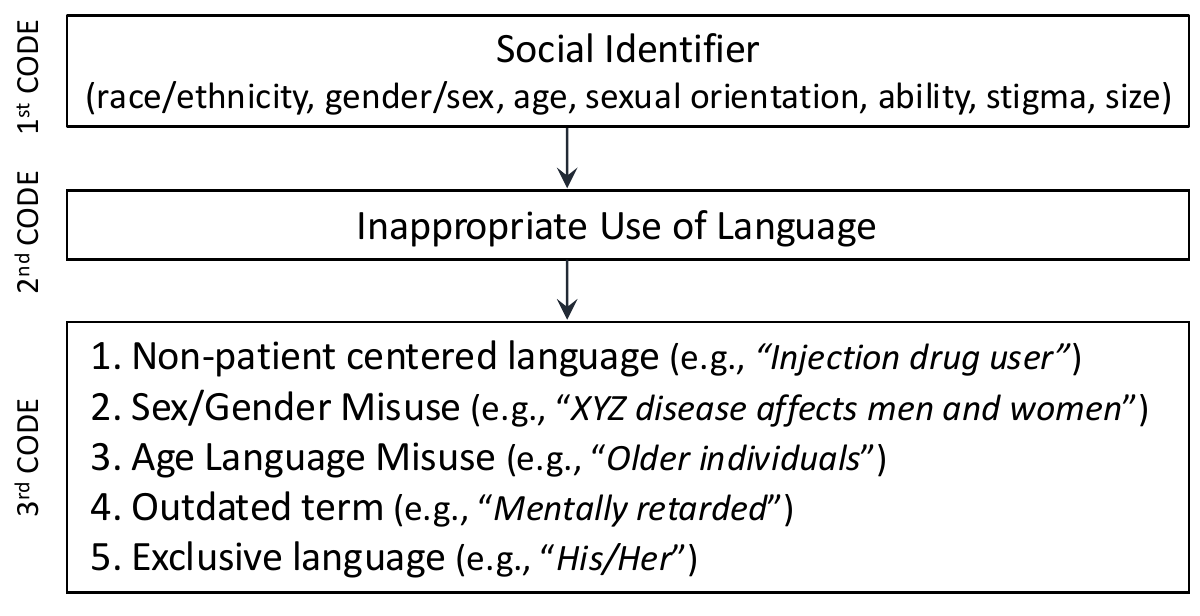} % Adjust the width as necessary
\caption{
% \fm{Recreate this figure from scratch just showing the leftmost branch.} 
\bricc{} Coding procedure for IUL. Annotators applied a structured 3-level coding process. The \textbf{1$^\textrm{st}$ code} identifies the presence of a social identifier (e.g., race, gender/sex, age). The \textbf{2$^\textrm{nd}$ code} flags the excerpt for IUL. The \textbf{3$^\textrm{rd}$ code} specifies the IUL subcategory. %This multi-level structure allows annotators to precisely capture how inappropriate or stigmatizing language manifests in clinical education materials.}
}

  \label{fig:bricc}
\end{figure}

Each annotated excerpt is structured within a three-level hierarchical coding scheme (Figure~\ref{fig:bricc}), as detailed in~\citet{salavati2024}. For the purposes of this work, we focus specifically on the levels that capture \textbf{Inappropriate Use of Language}, defined as:

\begin{quote}
``\textit{The use of inappropriate language to describe social identities. `Inappropriate use of language' refers only to the way a claim is described.}''
\end{quote}

This definition underscores that IUL is determined not by the factual accuracy or intent of a statement, but by the linguistic framing of social identities. In contrast to broader notions of bias that address content or consequences, IUL specifically targets language that is outdated, stigmatizing, exclusionary, or inconsistent with contemporary standards of respectful clinical communication.

\begin{figure*}[htbp]
    \centering \includegraphics[width=0.9\linewidth]{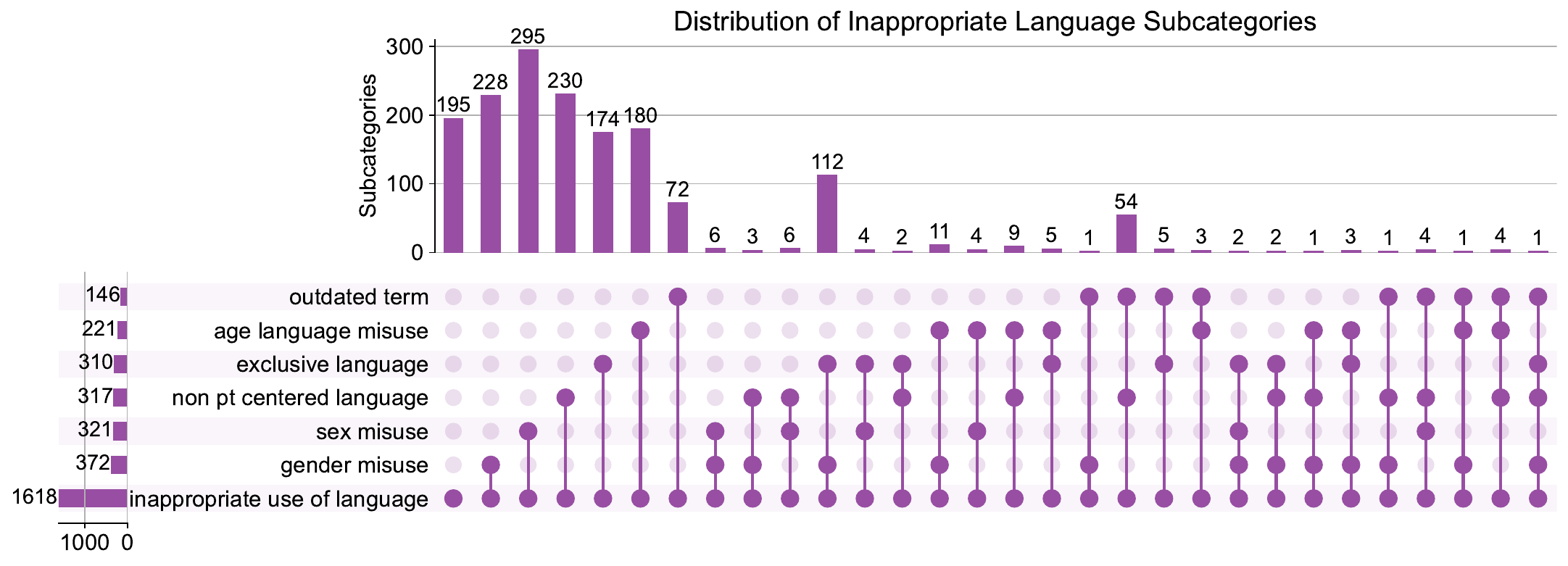}
    \caption{ Histogram illustrating the intersections among sets of IUL quotes, where filled circles represent the inclusion of specific IUL subcategories.}
    \label{fig:upset}
\end{figure*}

 Tables~\ref{tab:IUL_Samples} and~\ref{tab:Non_IUL_Samples} provide representative examples of excerpts labeled as IUL and non-IUL, respectively. Furthermore, Figure~\ref{fig:upset} visualizes the distribution of IUL subcategories and highlights the frequency of their co-occurrence. Notably, the most common overlaps are observed between \textbf{gender misuse} and \textbf{exclusive language} and between \textbf{non--patient-centered language} and \textbf{outdated terminology}, suggesting that modeling cross-category relationships may enhance fine-grained IUL classification. 
 
 In this dataset, there is substantial overlap between samples that do not contain IUL and those that exhibit potential bias. In many cases, it is difficult to clearly distinguish between the two, and these annotated negatives for IUL are particularly valuable for training the models. For example, in Table~\ref{tab:Non_IUL_Samples}, the sample for sex misuse---``\textit{Numerous measures of sexual function change as males age, including a decline in the frequency of orgasms, an increase in erectile dysfunction (ED), and a decline in the quality and quantity of sexual thoughts and enjoyment.}''---reflect potential sex bias, but it does not meet the criteria for sex misuse.  

\section{Methodology}
 Our methodology includes five steps: data preprocessing, positive and negative sampling, stratified multi-label data splitting, and model-based classification. Figure~\ref{fig:overview} illustrates the overall pipeline for detecting IUL in medical documents.

\subsubsection{Data Preprocessing.}
Many annotated excerpts were short sentence fragments with low word counts, making them difficult to interpret—even for trained experts—without surrounding context. To ensure interpretability for downstream tasks, we excluded quotes with fewer than four words. The remaining excerpts were cleaned by collapsing excess whitespace and removing leading or trailing punctuation. 

Because documents were independently assigned to multiple annotators, a single excerpt could receive different annotation spans or sets of codes. To consolidate overlapping or related samples, we defined a \textit{group of related quotes} \( G \) as any maximal subset of quotes such that for every \( x \in G \), there exists a \( x' \in G \) where \( x \) is a substring of \( x' \) or \( x = x' \). Within each group, we kept the longest quote $x^\star$ and merged all associated annotation codes by computing the union:
\begin{equation}
\ell(x^\star) = \bigcup_{x' \in G} \ell(x'),
\end{equation}
where \( \ell(.) \) denotes the set of annotation codes for a quote. This strategy served to establish consensus labels as well as to prevent data snooping by ensuring that fragments from the same original document were not split across training and test sets.

In this study, we focus on the set $\mathcal{C}$ of six most frequent subcategories of IUL (third level of the annotation schema), whose definitions are shown in Table~\ref{tab:IUL_Samples} alongside example quotes and the corresponding annotator comment. Specifically, $\mathcal{C} = $\{gender misuse, sex misuse, age-language misuse, non--patient-centered language, outdated terminology\}.%, as identified in the third level of our annotation schema:

\subsubsection{Positive Filter.}  
This filter is used for selecting the positive samples. %For the positive filters, we selected the subset of excerpts that we consider to exhibit inappropriate use of language, based on their assigned annotation codes.
Let $x$ be a text excerpt and $\ell(x)$ be the corresponding set of codes assigned by the annotators. We define the general IUL positive label for $x$ as:
\begin{equation}
y_\textrm{IUL} = \begin{cases}
1 & \textrm{if } \textrm{IUL} \in \ell(x), \textrm{ and } |\mathcal{C} \cap \ell(x)|\geq 1,\\
0 & \textrm{otherwise},
\end{cases}
\end{equation}
where $\mathcal{C}$ is the set of codes that represent the IUL subcategories that we are interested in studying. This captures whether the excerpt was flagged for any form of IUL.

We further define subcategory-specific IUL labels as:
\begin{equation}
y_c =
\begin{cases}
1 & \text{if } y_\text{IUL} = 1 \ \text{and} \ c \in \ell(x), \\
0 & \text{otherwise}.
\end{cases}
\end{equation}

\subsubsection{Negative Filter.}
To improve model robustness in distinguishing IUL from related biases, we construct two sets of non-trivial negative samples.
%To train a robust model capable of distinguishing between IUL and other related forms of potential bias, we construct two distinct sets of challenging negative samples.

\textit{Annotated Negatives (AN).} This set consists of excerpts explicitly coded by experts which do not include IUL labels but that contain social identifiers and/or express potential bias (e.g., related to dimensions such as gender, sex, or age). These samples are included to help the model learn to differentiate between bias and true IUL cases, ensuring it does not mistakenly flag bias-only examples as IUL.

\textit{Extracted Negatives (EN).} This set is derived from the larger corpus of medical instructional materials that produced the \bricc{} dataset. All EN samples contain at least one social identifier (SI) related to sex, gender, age, exclusive language, or outdated or non–patient-centered terminology. For age-related misuse, we also include excerpts matching regex patterns (e.g., ``65 year old''). To construct this set, we curated comprehensive SI lists from multiple sources, including the annotators’ detailed coding manual, which contains hundreds of terms specific to each IUL subcategory.

  These hard negatives are particularly challenging because, although they are labeled as appropriate, they often closely resemble positive samples in structure or tone. This setup allows for a rigorous evaluation of each model's discriminative ability. %\ch{This setup allows us to clearly test how well each model can tell the difference between appropriate and inappropriate language.}

Together, the AN and EN sets provide the model with challenging negative cases, enabling it to better discriminate between sentences that merely mention sensitive topics and those that actually reflect IUL.

\paragraph{Stratified Multi-label Splitting} 
Since this is a multilabel dataset, instead of a traditional stratified split, we applied a multilabel stratified split in two stages. First, we used \textit{Multilabel Stratified K-Fold} (MSKF) \cite{sechidis2011stratification} to divide the dataset into outer folds, ensuring that the distribution of labels is preserved across the training+validation and test sets. Then, within each training+validation partition, we used \textit{Multilabel Stratified Shuffle Split} (MLSSS) \cite{szymanski2017network} to further split the data into training and validation subsets in a randomized yet label-stratified manner. This two-stage strategy ensures that, in each fold:
\[
\Pr(\ell \in \ell(x_{\text{train}})) \approx \Pr(\ell \in \ell(x_{\text{val}})) \approx \Pr(\ell \in \ell(x_{\text{test}}))
\]
for each label $\ell \in \{y, z_1, \dots, z_C\}$. Using MSKF at the outer level ensures balanced evaluation across folds, while MLSSS within each fold introduces variability and reduces the risk of overfitting during model selection. An overview of the complete data curation, labeling, and splitting process is provided in Figure~\ref{fig:overview} (top).

\paragraph{Modeling Approaches} 
We propose four complementary strategies for detecting IUL in medical documents and evaluate them on the different
tasks, as illustrated in the bottom part of Figure~\ref{fig:overview}. These include: (I) a binary IUL classifier for detecting the general presence of IUL; (II) a multilabel classifier for simultaneously predicting multiple IUL subcategories; (III) a two-stage hierarchical classifier that first identifies whether a sample contains IUL and, if so, classifies its specific subcategory; and (IV) a prompting-based approach using large language models (LLMs), where each model receives tailored prompts and few-shot examples to predict labels for an unseen test set. While the primary focus is on supervised models (I–III), the LLM-based evaluation (IV) offers insights into the few-shot generalization capabilities of foundation models in this domain.

\subsubsection{General IUL Detection.}

The first step in our pipeline is to detect whether a given medical text excerpt contains an instance of IUL. We formulate this task as a supervised binary classification problem. Given a text input $x \in \mathcal{X}$, the goal is to predict a binary label $y \in \{0, 1\}$, where $y = 1$ indicates the presence of IUL and $y = 0$ indicates the converse.

We fine-tune a transformer-based classifier $f_\theta: \mathcal{X} \rightarrow [0,1]$ (state-of-the-art transformer-based architectures include DistilBERT \cite{sanh2019distilbert} and BioBERT \cite{lee2020biobert}, which are well-suited for handling domain-specific linguistic nuances in biomedical and clinical texts), parameterized by $\theta$, to estimate the probability $f_\theta(x)$ that the input contains IUL. We experiment with both DistilBERT ({distilbert-base-uncased}) and BioBERT (dmis-lab/biobert-base-cased-v1.1) as the underlying encoders. The predicted label is then computed as:
\[
\hat{y} = \begin{cases}
1 & \text{if } f_\theta(x) > 0.5, \\
0 & \text{otherwise}.
\end{cases}
\]

The model is trained using the weighted binary cross-entropy loss to address class imbalance:
\[
\mathcal{L}_\theta = - w_1 \cdot y \log f_\theta(x) - w_0 \cdot (1 - y) \log(1 - f_\theta(x)),
\]
where $w_0$ and $w_1$ are class weights inversely proportional to the frequences of class 0 and 1, respectively.

This general IUL classifier  serves as the first-stage filter in our hierarchical classification pipeline, routing positively predicted samples to downstream subcategory classifiers.

\subsubsection{Specific IUL Detection.}

Following the same principles as in the general IUL detection, we train six independent binary classifiers to detect each of the following IUL subcategories: \textit{gender misuse}, \textit{sex misuse}, \textit{age-related language misuse}, \textit{exclusive language}, \textit{non-patient-centered language}, or \textit{outdated terminology}.
For each subcategory $c \in \mathcal{C}$, we define a separate supervised binary classification problem. Given a text $x \in \mathcal{X}$, the task is to predict a binary label $z_c \in \{0,1\}$, where $z_c=1$ indicates that the text contains IUL of type $c$.

Each classifier $f_\theta^{(c)}: \mathcal{X} \rightarrow [0,1]$ outputs the probability that the input belongs to subcategory $c$. The predicted label is assigned as:
\[
\hat{z}_c = 
\begin{cases}
1 & \text{if } f_\theta^{(c)}(x) > 0.5, \\
0 & \text{otherwise}.
\end{cases}
\]

For each subcategory $c$, a sample is labeled as positive ($z_c=1$) if it is annotated with the corresponding IUL subcategory. Negative samples ($z_c=0$) are selected among those (i) not labeled as IUL and that (ii) also contain relevant social identifiers or, in the case of age-related language misuse, match an age expression pattern (e.g., "65 year old").

Each subcategory classifier fine-tunes a separate instance of the DistilBERT model. 

Training minimizes a weighted binary cross-entropy loss:
\[
\mathcal{L}^{(c)}_\theta = - w_1^{(c)} z_c \log f_\theta^{(c)}(x) - w_0^{(c)} (1 - z_c) \log(1 - f_\theta^{(c)}(x)),
\]
where $w_0^{(c)}$ and $w_1^{(c)}$ are class weights computed based on the distribution of positive and negative samples for $c$.

\subsubsection{Multilabel IUL Detection.}

As a third approach, we formulate IUL detection as a multilabel classification task, where a single text excerpt can simultaneously belong to multiple IUL subcategories. Given a text input $x \in \mathcal{X}$, the goal is to predict a label vector $\mathbf{z} = (z_0,z_1, z_2, \dots, z_{C}) \in \{0,1\}^C$, where $z_0$ indicates whether or not $x$ is predicted as non-IUL whereas the following $C$ elements are subcategory-specific predictions. Outputting $z_0$ (as part of a flat hierarchical inference) avoids the need to combine subcategories' predictions post hoc during general IUL detection.

We fine-tune a single DistilBERT-based transformer classifier $f_\theta: \mathcal{X} \rightarrow [0,1]^{C+1}$, where each output $\hat{z}_c = f_\theta(x)_c$ represents the predicted probability that input $x$ belongs to subcategory $c$. Binary predictions are assigned per subcategory using:
\[
\hat{z}_c = 
\begin{cases}
1 & \text{if } f_\theta(x)_c > 0.5, \\
0 & \text{otherwise}.
\end{cases} \quad \textrm{for } c=1,\ldots,C
\]

We apply the binary cross-entropy loss across all labels:
\[
\mathcal{L}_{\theta} = - \sum_{c=1}^C  z_c \log f_\theta(x)_c + (1 - z_c) \log(1 - f_\theta(x)_c) .
\]
This joint formulation allows the model to capture dependencies and co-occurrence patterns between subcategories, providing a more holistic view of IUL signals.

\begin{figure*}[htbp]
    \centering \includegraphics[width=0.98\linewidth]{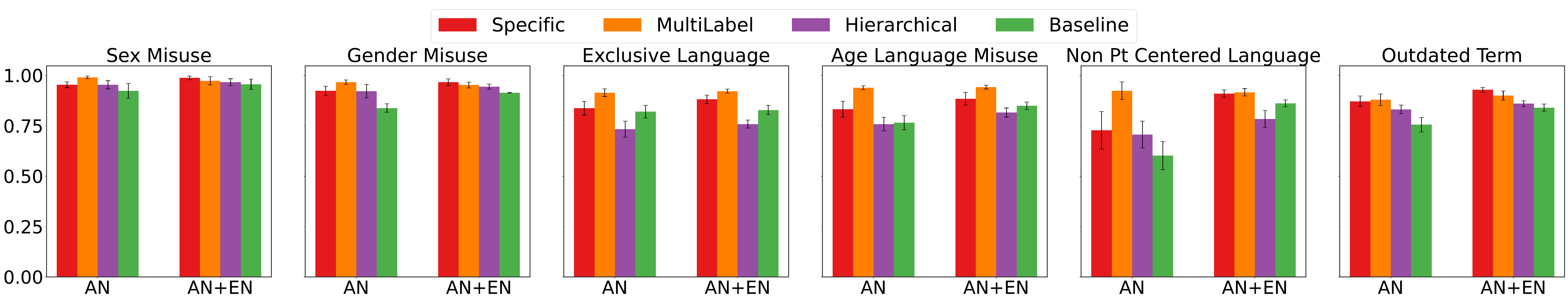}
    \caption{AUC performance of Specific, Multilabel, Hierarchical, and baseline models for each IUL subcategory when trained on AN versus when trained on AN+EN.}
    \label{fig:AUCbars}
\end{figure*}
\subsubsection{Two-stage Hierarchical IUL Classification.}
We implement a hierarchical classification pipeline following the \textit{Local Classifier Per Node} approach \cite{silla2011survey}. In this strategy, a general classifier first determines whether a sample belongs to the broad IUL category. If so, the sample is passed to specialized subcategory classifiers, each trained independently. This top-down, modular design leverages the hierarchical structure of the task, enabling localized decision boundaries that improve both interpretability and performance over flat classification methods.

Given an input text $x \in \mathcal{X}$, the hierarchical decision process consists of two stages:
\begin{itemize}
    \item \textbf{Level 1 (General Detection):} A binary classifier $f_\theta: \mathcal{X} \rightarrow [0,1]$ predicts whether the text contains any form of IUL as $\hat{y} = \mathds{1}\{f_\theta(x) > 0.5\}$.
    \item \textbf{Level 2 (Subcategory Detection):} For texts predicted as containing IUL ($\hat{y}=1$), a multilabel classifier $f_\theta: \mathcal{X} \rightarrow [0,1]^C$ predicts each of $C$ specific IUL subcategories as $\hat{\mathbf{z}}$, where $\hat{z}_c = \mathds{1}\{f_\theta(x)_c > 0.5\}$.
\end{itemize}

This is a combination of the general and multilabel classifiers described earlier. For the multilabel part, we only used our six IUL subcategories as labels.

\subsubsection{Few-Shot Prompting for General Classification} We experiment with different LLMs, namely MetaLlama-3.1 8B, MetaLlama-3.3 70B \cite{grattafiori2024llama}, MedLlama3 8B, and GPT-4o \cite{islam2024gpt}. We test various prompt formulations, including those containing definitions of IUL subcategories, curated examples of IUL, and combinations of both. Details of the prompt templates and evaluation setup are provided in the Appendix A\ref{appendix:prompts}. %\fm{fixed!} \ch{why A does not show up here?.}
Each of our experiments classifies text excerpts as either positive or negative for IUL.

\begin{table*}[tbp]
\centering
\resizebox{\textwidth}{!}{%
\begin{tabular}{l|ccc|ccc|ccc|ccc|ccc}
\toprule
\multirow{2}{*}{\textbf{Method}} & \multicolumn{3}{c|}{\textbf{\underline{Precision}}} & \multicolumn{3}{c|}{\textbf{\underline{Recall}}} & \multicolumn{3}{c|}{\textbf{\underline{F1-Score}}} & \multicolumn{3}{c|}{\textbf{\underline{F2-Score}}} & \multicolumn{3}{c}{\textbf{\underline{AUC}}} \\ 
              & AN   & EN    & AN+EN   & AN   & EN    & AN+EN   & AN   & EN    & AN+EN   & AN   & EN    & AN+EN   & AN   & EN    & AN+EN \\
\midrule
General (DistilBert)  & .473 & .539 & .652 & \textbf{.792} & .885 & .846 & .584 & .668 & \textbf{.735} & .690 & \textbf{.783} & .797 & .870 & .919 & \textbf{.943} \\
General (BioBert)     & .443 & .498 & .568 & .760 & \textbf{.902} & \textbf{.898} & .555 & .641 & .694 & .660 & .775 & \textbf{.803} & .848 & .899 & .940 \\
MultiLabel            & \textbf{.795} & \textbf{.818} & \textbf{.786} & .681 & .658 & .670 & \textbf{.731} & \textbf{.725} & .717 & \textbf{.700} & .683 & .687 & \textbf{.941} & \textbf{.944} & .941 \\
Baseline             & .294 & .581 & .742 & .754 & .542 & .311 & .423 & .560 & .438 & .574 & .549 & .352 & .755 & .863 & .865 \\ 
\bottomrule
\end{tabular}
}
\caption{Performance of general IUL detection trained on different sets of negative examples (AN: annotated negatives, EN: extracted negatives, AN+EN: combined negatives).}
\label{tab:general_IUL}
\end{table*}

\begin{table}[tbp]
\centering
\small
\begin{tabular}{lrrrrr}
\toprule
\textbf{Model} & \textbf{\underline{P}} & \textbf{\underline{R}} & \textbf{\underline{F1}} & \textbf{\underline{F2}} & \textbf{\underline{AUC}}
\\
\midrule
MedLlama3 8B & .163 & .991 & .280 & .491 & .491\\
Llama3.1-8B/3.3-70B & \textbf{.179} & \textbf{1.000} & \textbf{.304} & \textbf{.522} & .523/.500\\
%Llama3.1 70B & .179 & \textbf{1.000} & .304 & .522 & .500\\
GPT-4o & .156 & .801 & .261 & .439 & \textbf{.550} \\
\bottomrule
\end{tabular}
\caption{Performance comparison of \textsc{MedLlama3-8B}, \textsc{MetaLlama3.1-8B/3.3-70B}, and GPT-4o on general IUL detection using few-shot prompting across AN+EN.}
\label{tab:llm_few_shot}
\end{table}

\begin{table}[h!]
\centering
\begin{tabular}{clccccc}
    \toprule
    & \textbf{Method} & \textbf{\underline{P}} & \textbf{\underline{R}} & \textbf{\underline{F1}} & \textbf{\underline{F2}} & \textbf{\underline{AUC}} \\
    \midrule
    \parbox[t]{2mm}{\multirow{4}{*}{\rotatebox[origin=c]{90}{Gender}}} 
    & Specific    & .454 & \textbf{.863} & .584 & .717 & .924 \\ 
    & MultiLabel     & \textbf{.816} & .795 & .803 & .798 & \textbf{.973} \\
    & Hierarchical   & .763 & .857 & \textbf{.805} & \textbf{.835} & .921 \\
    & Baseline       & .663 & .712 & .687 & .702 & .884 \\
    \midrule
    \parbox[t]{2mm}{\multirow{4}{*}{\rotatebox[origin=c]{90}{Sex}}} 
    & Specific    & .452 & \textbf{.897} & .590 & .736 & .954 \\ 
    & MultiLabel     & \textbf{.912} & \textbf{.897} & \textbf{.903} & \textbf{.899} & \textbf{.994} \\
    & Hierarchical   & .870 & .875 & .868 & .871 & .953 \\
    & Baseline       & .635 & .625 & .630 & .627 & .854 \\
    \midrule
    \parbox[t]{2mm}{\multirow{4}{*}{\rotatebox[origin=c]{90}{Age}}} 
    & Specific    & .236 & \textbf{.747} & .355 & \textbf{.513} & .833 \\ 
    & MultiLabel     & \textbf{.544} & .412 & \textbf{.465} & .431 & \textbf{.911} \\
    & Hierarchical   & .370 & .538 & .434 & .489 & .758 \\
    & Baseline       & .333 & .523 & .407 & .469 & .789 \\
    \midrule
    \parbox[t]{2mm}{\multirow{4}{*}{\rotatebox[origin=c]{90}{Exc-Lang}}} 
    & Specific    & .660 & .742 & .694 & \textbf{.721} & .837 \\ 
    & MultiLabel     & \textbf{.762} & .223 & .322 & .253 & .775 \\
    & Hierarchical   & .670 & .532 & .578 & .547 & .734 \\
    & Baseline       & .667 & \textbf{.903} & \textbf{.767} & .244 & \textbf{.867} \\
    \midrule
    \parbox[t]{2mm}{\multirow{4}{*}{\rotatebox[origin=c]{90}{Non-Pt}}} 
    & Specific    & .635 & .968 & \textbf{.767} & .876 & .727 \\ 
    & MultiLabel     & \textbf{.888} & .599 & .713 & .639 & \textbf{.819} \\
    & Hierarchical   & .757 & .631 & .685 & .651 & .707 \\
    & Baseline       & .606 & \textbf{1.000} & .754 & \textbf{.885} & .616 \\
    \midrule
    \parbox[t]{2mm}{\multirow{4}{*}{\rotatebox[origin=c]{90}{Outdated}}} 
    & Specific    & .404 & \textbf{.802} & .533 & \textbf{.664} & .871 \\ 
    & MultiLabel     & \textbf{.894} & .288 & .435 & .333 & \textbf{.878} \\
    & Hierarchical   & .721 & .452 & \textbf{.555} & .488 & .832 \\
    & Baseline       & .368 & .724 & .488 & .607 & .768 \\
    \bottomrule
\end{tabular}
\caption{
% \fm{Maybe include a column showning the rank of each model for each subcategory based on AUC?} 
Performance of Binary (type-specific), Multilabel, Two-stage Hierarchical, and Baseline models trained on \textbf{AN} on detection of each IUL type. 
% Metrics: \underline{P}recision, \underline{R}ecall, \underline{F1}-score, \underline{F2}-score, \underline{AUC}.
}
\label{tab:results_an}
\end{table}

\begin{table}[h!]
\centering
\begin{tabular}{clccccc}
    \toprule
    & \textbf{Method} & \textbf{\underline{P}} & \textbf{\underline{R}} & \textbf{\underline{F1}} & \textbf{\underline{F2}} & \textbf{\underline{AUC}} \\
    \midrule
    \parbox[t]{2mm}{\multirow{4}{*}{\rotatebox[origin=c]{90}{Gender}}} 
    & Specific    & .759 & \textbf{.876} & .809 & .847 & .966 \\ 
    & MultiLabel     & \textbf{.820} & .777 & .794 & .782 & \textbf{.969} \\
    & Hierarchical   & .765 & .879 & \textbf{.817} & \textbf{.853} & .944 \\
    & Baseline       & .818 & .347 & .487 & .392 & .914 \\
    \midrule
    \parbox[t]{2mm}{\multirow{4}{*}{\rotatebox[origin=c]{90}{Sex}}} 
    & Specific    & .837 & .904 & .865 & .887 & \textbf{.988} \\ 
    & MultiLabel     & \textbf{.929} & .878 & \textbf{.902} & \textbf{.887} & \textbf{.988} \\
    & Hierarchical   & .874 & \textbf{.882} & .876 & .879 & .966 \\
    & Baseline       & .813 & .583 & .677 & .617 & .955 \\
    \midrule
    \parbox[t]{2mm}{\multirow{4}{*}{\rotatebox[origin=c]{90}{Age}}} 
    & Specific    & .379 & \textbf{.639} & .452 & \textbf{.538} & .884 \\ 
    & MultiLabel     & .537 & .462 & .464 & .457 & \textbf{.909} \\
    & Hierarchical   & .455 & .616 & \textbf{.517} & .569 & .815 \\
    & Baseline       & \textbf{.571} & .118 & .194 & .140 & .849 \\
    \midrule
    \parbox[t]{2mm}{\multirow{4}{*}{\rotatebox[origin=c]{90}{Exc-Lang}}} 
    & Specific    & .663 & \textbf{.852} & \textbf{.742} & \textbf{.803} & \textbf{.881} \\ 
    & MultiLabel     & \textbf{.735} & .216 & .323 & .249 & .797 \\
    & Hierarchical   & .627 & .652 & .635 & .644 & .759 \\
    & Baseline       & .721 & .613 & .662 & .632 & .828 \\
    \midrule
    \parbox[t]{2mm}{\multirow{4}{*}{\rotatebox[origin=c]{90}{Non-pt}}} 
    & Specific    & .849 & \textbf{.893} & \textbf{.868} & \textbf{.882} & \textbf{.910} \\ 
    & MultiLabel     & \textbf{.895} & .542 & .670 & .586 & .842 \\
    & Hierarchical   & .810 & .643 & .709 & .667 & .784 \\
    & Baseline       & .800 & .874 & .834 & .858 & .861 \\
    \midrule
    \parbox[t]{2mm}{\multirow{4}{*}{\rotatebox[origin=c]{90}{Outdated}}} 
    & Specific    & .602 & \textbf{.775} & \textbf{.666} & \textbf{.724} & \textbf{.929} \\ 
    & MultiLabel     & \textbf{.846} & .247 & .379 & .287 & .918 \\
    & Hierarchical   & .692 & .501 & .574 & .527 & .860 \\
    & Baseline       & .758 & .233 & .355 & .270 & .840 \\
    \bottomrule
\end{tabular}
\caption{Performance of binary (type-specific), multilabel, two-stage hierarchical, and baseline models trained on \textbf{AN+EN} on detection of each IUL type. 
% Metrics: \underline{P}recision, \underline{R}ecall, \underline{F1}-score, \underline{F2}-score, \underline{AUC}. Best result for each metric is bold-faced.
}
\label{tab:results_all}
\end{table}
\section{Experimental Setup}
We adopt a unified experimental framework across all four IUL detection strategies to ensure consistent evaluation and fair comparison. All supervised models are trained independently using MSKF and MLSSS ($k=5$), with each fold comprising separate training, validation, and test sets that preserve class distributions across both general IUL and subcategory-specific labels. For all experiments, we evaluate filtered variants using AN and EN in the training data.

Transformer-based models are fine-tuned using either DistilBERT (distilbert-base-uncased) or BioBERT. Input texts are tokenized and padded or truncated to a maximum length of 512 tokens. Training is performed using the AdamW optimizer with a learning rate of $4\times10^{-5}$, batch size of 32, and early stopping with a patience of 10 epochs.

As a non-transformer baseline, we train an XGBoost classifier using BioWordVec embeddings pretrained on PubMed and MIMIC-III~\cite{chen2019}. These baseline models are evaluated on both general IUL and subcategory classification tasks. While transformer-based models use fixed hyperparameters, the XGBoost baseline undergoes hyperparameter tuning via Optuna~\cite{optuna_2019} with early stopping, using the same evaluation metrics for the sake of comparison.

We evaluate all models using standard metrics: precision, recall, F1 score, F2 score, area under the ROC curve (AUC), and the confusion matrix. Reported results are averaged across five cross-validation folds. For comparison with general IUL models, we derive a binary prediction from the multilabel classifier by taking the maximum predicted subcategory probability and applying a 0.5 threshold.

To evaluate foundation models, we use few-shot prompting with LLMs such as GPT-4o and LLaMA variants. Each model receives a small set of labeled examples alongside an unseen test instance and is prompted to determine whether the input contains IUL.

\section{Results} In this section, we present the performance results of four modeling approaches for IUL detection, comparing multiple architectures and training configurations to identify the most robust strategies across key evaluation metrics.

\begin{figure}[tbp] 
    \centering
    \includegraphics[width=\columnwidth]{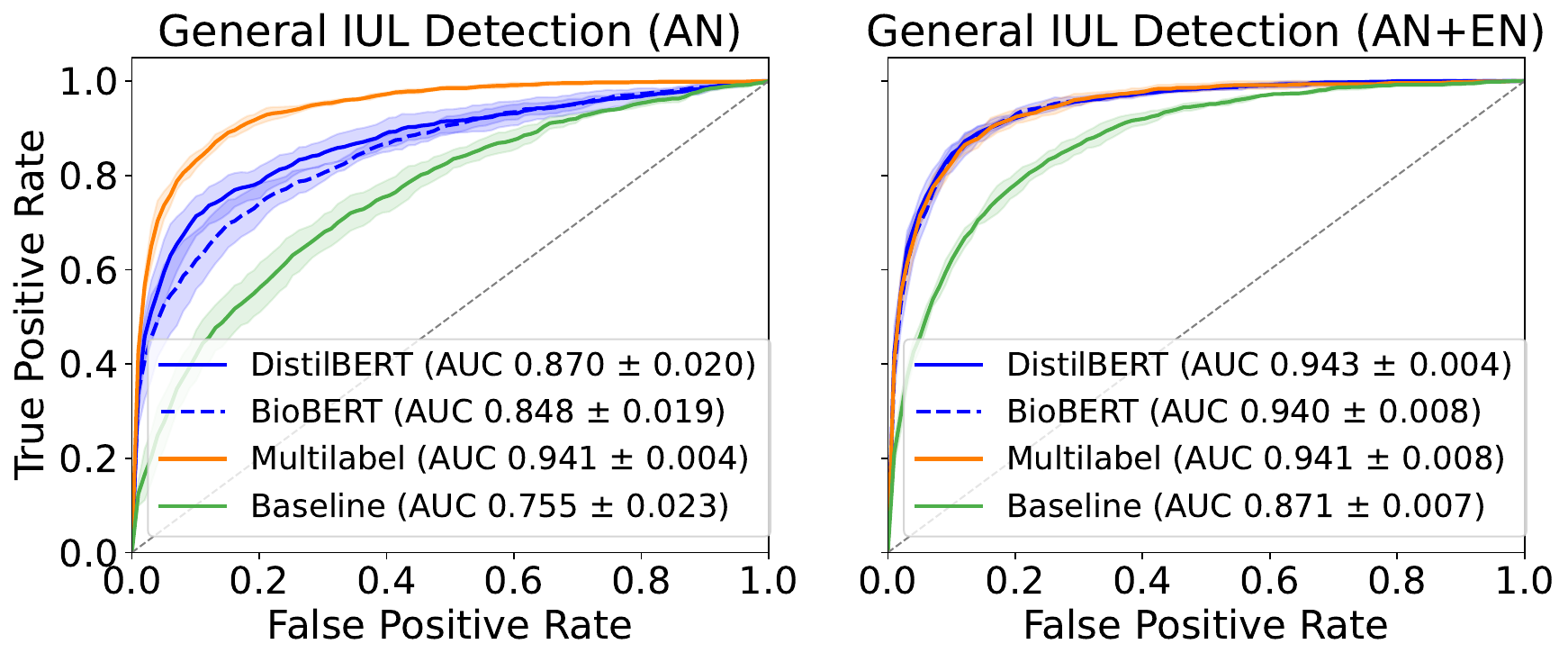} % Adjust the width as necessary
\caption{AUC plots for general IUL detection trained on different sets of negatives (AN vs.\ AN+EN).}
  \label{fig:AUCplot}
\end{figure}

\paragraph{General IUL Detection.}
Table \ref{tab:general_IUL} summarizes the performance results on general IUL detection for different modeling strategies (General-DistilBERT, General-BioBERT, Multilabel, and Baseline) and training setups using varying negative sets (AN, EN, and AN+EN). The Multilabel model consistently yields the highest precision and strong F1-scores across all settings, reflecting its robustness in reducing false positives and balancing precision-recall. BioBERT achieves the highest recall, especially with EN, highlighting its sensitivity to true positives. While AUC for Multilabel remains stable, both DistilBERT and BioBERT benefit from AN+EN, showing improved AUC and F2-scores. Overall, the Multilabel model demonstrates reliable and competitive performance across metrics and configurations.

Figure~\ref{fig:AUCplot} shows AUC results for general IUL detection, comparing models trained on AN-only negatives versus those trained on the combined AN+EN set. When trained on AN alone, the Multilabel model substantially outperforms the Baseline. However, this performance gap narrows significantly with the inclusion of EN samples, and the difference between the General and Multilabel models nearly disappears. These findings suggest that EN samples provide more informative and challenging negative examples, helping to equalize performance across models.

Table~\ref{tab:llm_few_shot} shows the LLMs performance on IUL detection. Among the evaluated models, LLaMA variants achieved perfect recall but low precision, likely due to a strong bias toward predicting the positive class, resulting in AUC scores near random. The LLaMA3 8B and 70B models performed nearly identically, with the 8B model slightly outperforming in AUC--indicating that larger model size offers little benefit under current settings. MedLLaMA3 also reached near-perfect recall with similarly low precision. GPT-4o was outperformed by all fine-tuned LLaMA variants except in AUC which is comparable. Overall, LLaMA3.1 8B emerged as the most effective and cost-efficient LLM for high-recall IUL detection, though its performance still lagged behind that of the SLM-based models.

\paragraph{IUL Subcategory Detection.} 
Table~\ref{tab:results_an} presents results on annotated negative samples (AN), which were carefully assessed by experts for each IUL subcategory. Overall, the Multilabel model achieves the highest AUC scores across most categories, demonstrating strong generalization and robustness. The only exception is the Exclusive Language subcategory, where the Baseline model outperforms others. This likely reflects the nature of Exclusive Language, which often depends on the presence or absence of specific lexical items. The Baseline model relies on static, word-level features that are suited for detecting such surface-level patterns. In contrast, transformer-based models like BERT prioritize contextual semantics and may struggle when fine-grained lexical cues are key. These findings suggest that while transformer models excel at semantic understanding, traditional embedding-based models may retain an edge for lexically grounded tasks like exclusive language detection.

Table \ref{tab:results_all} shows, in turn, the performance results for IUL subcategory classification when the training data also includes Extracted Negatives (AN+EN). Similarly to the case of general IUL detection, including EN yields large performance gains for most models. Specifically for AUC, the only approach that does not always benefit from adding Extracted Negatives is the Multilabel classifier (recall that this is the best performing model when trained on AN only). In comparison to the results based on AN only, the AUC gap between the specific classifiers and multilabel is much smaller for Gender, Sex and Age, varying between 0.0 and 2.8\%. Moreover, specific classifiers actually outperform multilabel on Exclusive Language, Non-patient-centered Language and Outdated Terminology.

Figure \ref{fig:AUCbars} illustrates how training on AN+EN (which includes additional hard negatives) impacts the AUC performance across six IUL subcategories. In most cases, incorporating EN improves performance or narrows the gap between models. Notably, the Multilabel model shows substantial gains in challenging categories such as Non–patient-entered Language and Age Language Misuse.

\begin{figure}[tbp] 
    \centering
    \includegraphics[width=\columnwidth]{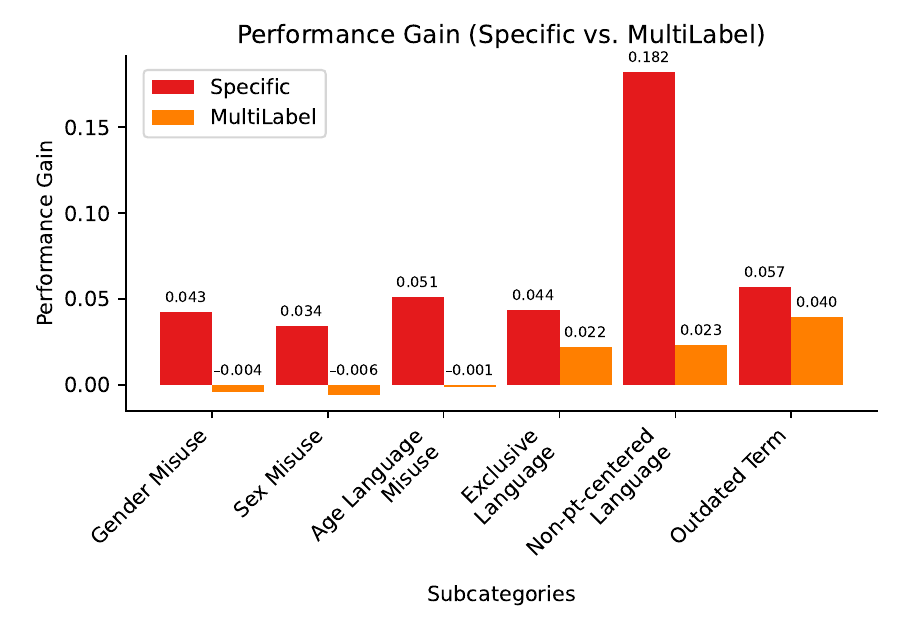} % Adjust the width as necessary
\caption{Performance gain from including EN in training specific and multilabel models across IUL subcategories. The gain is computed as the difference in AUC when models are trained on AN+EN versus AN only.}
  \label{fig:bar}
\end{figure}

To better investigate this phenomenon, we compute the AUC gain achieved by adding EN to the training data for both specific and multilabel models across different IUL subcategories, illustrated in Figure~\ref{fig:bar}. %The bars reflect the improvement in AUC when models are trained on the combined set (AN+EN) compared to using annotated negatives (AN) alone.
Notably, the performance gains for Multilabel models are consistently smaller than those observed for specific models across all subcategories. This trend indicates that multilabel models are less sensitive to the additional information provided by ENs, suggesting they generalize better even with fewer negative examples. Conversely, specific models significantly benefit from the inclusion of ENs---especially in the Non-patient-entered and Outdated Term subcategories---showing marked improvement in discriminative performance. This difference may stem from the nature of these categories: they often rely on more explicit or pattern-like signals (e.g., specific outdated terms or exclusive phrases), which simpler binary classifiers can effectively capture without needing cross-label interactions. Overall, the Multilabel approach benefits most when the category has richer, more varied examples and cross-task dependencies (as seen in age, sex, and gender), whereas binary models excel when detecting more rigid, pattern-based categories like exclusive language or outdated terms.

The key takeaway is that when ENs are available, specific models tend to outperform Multilabel models in most subcategories, with the exception of Age Language Misuse, where Multilabel remains competitive.

\section{Conclusions \& Future Work}
Our analysis of inappropriate use of language in medical education is motivated by the fact that linguistic choices in clinical training materials can profoundly shape physicians’ attitudes and prescribing behaviors. If left unaddressed, such language may inadvertently contribute to healthcare disparities. To the best of our knowledge, this work represents the first comprehensive AI-driven effort to detect IUL from medical curricula.

In this study, we developed and evaluated several SLMs and LLMs for identifying IUL in medical instructional materials, with a particular emphasis on maximizing recall. Prioritizing recall ensures that potentially harmful language is consistently flagged for expert review in the first stage of our proposed expert-in-the-loop framework, designed to support the highest levels of quality control in medical communication. Our results show that the multilabel classifier performs best when trained solely on the expert-annotated dataset. Nonetheless, when the training data is augmented with additional negative samples---text excerpts presumed to be appropriate---individual subcategory classifiers surpass the multilabel approach in performance. These findings suggest that while the multilabel model is effective in well-annotated settings, subcategory-specific classifiers offer greater robustness and generalizability in more diverse, real-world contexts. Notably, we observe that while LLMs achieve high recall, their precision remains low and they are significantly outperformed by SLMs, underscoring the importance of model selection based on task-specific constraints and deployment goals.

The overarching objective of this work is to build automated systems that assist institutions in fostering more equitable and patient-centered care by analyzing, generating, and improving clinical text---including progress notes, diagnostic impressions, and other written communication. While promising, our current approach has limitations. We do not yet account for the intersectionality of bias and inappropriate use of language. Moreover, the models do not provide explanations for their classifications, which could improve transparency and user trust. Future work should explore integrating explainable AI techniques, refining precision-recall trade-offs based on clinical context, and evaluating system performance in real-world expert-in-the-loop settings.

% % \newpage
% \section*{Acknowledgements} This work supported in part by the National Science Foundation Grant IIS-2147305.% and the National Board of Medical Examiners Stemmler Fund 2021-3132. 

% \bigskip

\appendix
\section{Appendix A: Prompts for Zero-Shot and Few-Shot Experiments}
\label{appendix:prompts}

We designed three prompt types: (1) IUL category definitions, (2) annotated examples of IUL categories, and (3) both combined---the third was most effective.

\definecolor{appendixpurple}{RGB}{152, 78, 163}
\begin{tcolorbox}[breakable,colback=appendixpurple!10!white, colframe=appendixpurple!50!black, title=Prompt Template]
\begin{quote}
\textbf{System Prompt:} You are a clinical-language annotation assistant. Your task is to determine whether a given clinical text excerpt contains inappropriate use of language (IUL), regardless of whether the medical content is factually correct.

\textbf{Definitions of IUL Categories:}
\begin{itemize}
    \item \textbf{Gender Misuse}: Using gendered terms (e.g., ``women", ``men") where anatomical or sex-based terms are more accurate.
    \item \textbf{Sex Misuse}: Using terms like ``male" or ``female" to describe individuals, e.g., ``a 49-year-old male” instead of ``a 49-year-old man”.
    \item \textbf{Age Language Misuse}: Using vague or stigmatizing age terms like ``the elderly" or ``young people".
    \item \textbf{Exclusive Language}: Referring to binary groups like ``both males and females", which excludes non-binary individuals.
    \item \textbf{Non-Patient Centered Language}: Defining people by their conditions, e.g., ``diabetics” or ``alcoholics”, instead of ``patients with diabetes”.
    \item \textbf{Outdated Terms}: Using language no longer appropriate in clinical contexts, such as ``mentally retarded” or ``fat and fertile female”.
\end{itemize}

[Place examples here if included.]

\textbf{Task:} Consider the excerpt and identify any terms or patterns that might signal IUL based on the examples.

\textbf{Excerpt:} \\
Almost all women in the U.S ($>$ 99\%) who have sexual intercourse use contraception at some point during their reproductive life - including women of all races, nationalities and religions.

\textbf{Reasoning:} [Write a brief 1–2 sentence explanation.]

\textbf{Final Answer:}
\end{quote}
\end{tcolorbox}

\begin{tcolorbox}
[breakable,colback=appendixpurple!10!white, colframe=appendixpurple!50!black, title=Few-Shot Examples]   
\begin{quote}

\textbf{Examples:}
\begin{itemize}
    \item \textbf{Gender Misuse} \\
    \emph{Excerpt:} ``Often, significant changes in a child’s growth reflect significant events in the family unit such as a mother going to work, parents separating, moving to a new home or a significant family illness." \\
    \emph{Explanation:} This reinforces traditional family structures and may stigmatize families without a mother or where mothers work. However, this reflects gender bias rather than gender misuse. \\
    \emph{IUL Label:} 1

    \item \textbf{Sex Misuse} \\
    \emph{Excerpt:} ``Numerous measures of sexual function change as males age, including a decline in the frequency of orgasms, an increase in erectile dysfunction (ED), and a decline in the quality and quantity of sexual thoughts and enjoyment." \\
    \emph{Explanation:} This shows possible sex bias but does not misuse ``male” in the context of individuals. Suggest verifying accuracy. \\
    \emph{IUL Label:} 1

    \item \textbf{Age Language Misuse} \\
    \emph{Excerpt:} ``Hereditary pancreatitis (HP) is an autosomal dominant disease with 80\% penetrance, characterized by recurrent episodes of pancreatitis from childhood with a familial occurrence.'' \\
    \emph{Explanation:} ``Childhood” is appropriate in this context and is not vague. \\
    \emph{IUL Label:} 1

    \item \textbf{Exclusive Language} \\
    \emph{Excerpt:} ``The gross morphological appearance of the nuclear chromatin differs in cells between males and females." \\
    \emph{Explanation:} This reflects a binary framing of sex, but in this scientific context, it's acceptable.\\
    \emph{IUL Label:} 1

    \item \textbf{Non-Patient Centered Language} \\
    \emph{Excerpt:} ``A landmark study detailing the clinical features of alcoholic hepatitis; also, one of the first to demonstrate a potential benefit from corticosteroid therapy." \\
    \emph{Explanation:} ``Alcoholic” refers to the disease name here, not individuals. No IUL. \\
    \emph{IUL Label:} 1

    \item \textbf{Outdated Term} \\
    \emph{Excerpt:} ``Psychomotor retardation or agitation nearly every day that is observable by others." \\
    \emph{Explanation:} This is a proper clinical use of ``retardation” within a DSM context. Acceptable.\\
    \emph{IUL Label:} 1
\end{itemize}
\end{quote}
\end{tcolorbox}

\section*{Ethics Statement}
% We provide the following contextual information to address the ethical considerations of our project.
\paragraph{Ethical Considerations.} This study is part of a broader effort to improve medical education by identifying and addressing IUL within curricular materials. The tool we developed is designed to assist human reviewers, not to function autonomously. All flagged content should undergo human review to determine appropriateness in context. Our system is meant to support educators by drawing attention to potentially problematic language, fostering greater awareness and sensitivity in curriculum design. It remains the responsibility of medical institutions to take meaningful action in revising materials and providing faculty development that promotes respectful and inclusive language. We emphasize the importance of ensuring that the responsibility for improving language use is equitably shared and does not fall disproportionately on underrepresented faculty, to avoid contributing to the minority tax.

\paragraph{Adverse Impacts} This tool is not intended as a means to punish educators but as a constructive aid for improvement. Results should always be interpreted with care and contextual understanding. Automated detection of IUL is meant as an initial indicator for further human evaluation and should not be viewed as a definitive label. For instance, identifying higher frequencies of flagged language in a curriculum does not imply fault or intent on the part of faculty or students.

% \paragraph{Conflicts of Interest.} \fm{Are we obligated to include this section? This is also de-indetifying.} At the time of this research, Shiri Dori-Hacohen held a significant financial interest in AuCoDe.

\paragraph{Researcher Positionality.} Our team includes members from diverse disciplinary and personal backgrounds, spanning computer science, social science, and medicine. These varied perspectives have shaped our approach to IUL detection. We acknowledge that while this diversity offers valuable insight, it also brings inherent limitations.

We note that ChatGPT was used for light editing; full responsibility for the content remains with the authors.

% \newpage
\section*{Acknowledgements} This work supported in part by the National Science Foundation Grant IIS-2147305.% and the National Board of Medical Examiners Stemmler Fund 2021-3132. 

\bibliography{aaai25}

\end{document}